\documentclass{article}

\usepackage{graphicx,subfigure,natbib,tabularx,algorithm,algorithmic,units}

\usepackage[accepted]{whi2018}

\icmltitlerunning{Defining Locality for Surrogates in Post-hoc Interpretablity}

\begin{document} 

\twocolumn[
\icmltitle{Defining Locality for Surrogates in Post-hoc Interpretablity}

\icmlsetsymbol{equal}{*}

\begin{icmlauthorlist}
\icmlauthor{Thibault Laugel}{sor,equal}
\icmlauthor{Xavier Renard}{axa,equal}
\icmlauthor{Marie-Jeanne Lesot}{sor}
\icmlauthor{Christophe Marsala}{sor}
\icmlauthor{Marcin Detyniecki}{sor,axa,pol}
\end{icmlauthorlist}

\icmlaffiliation{sor}{Sorbonne Universit\'e, CNRS, LIP6, Paris, France}
\icmlaffiliation{axa}{AXA, Paris, France}
\icmlaffiliation{pol}{Polish Academy of Science, IBS PAN, Warsaw, Poland}

\icmlcorrespondingauthor{Thibault Laugel}{thibault.laugel@lip6.fr}
\icmlcorrespondingauthor{Xavier Renard}{xavier.renard@axa.com}

\icmlkeywords{interpretability, transparency}

\vskip 0.3in
]

\printAffiliationsAndNotice{\icmlEqualContribution}


\begin{abstract} 
Local surrogate models, to approximate the local decision boundary of a black-box classifier, constitute one approach to generate explanations for the rationale behind an individual prediction made by the back-box.
This paper highlights the importance of defining the right \emph{locality}, the neighborhood on which a local surrogate is trained, in order to approximate accurately the \emph{local} black-box decision boundary.
Unfortunately, as shown in this paper, this issue is not only a parameter or sampling distribution challenge and has a major impact on the \emph{relevance} and quality of the approximation of the local black-box decision boundary and thus on the \emph{meaning} and accuracy of the generated explanation. 
To overcome the identified problems, quantified with an adapted measure and procedure, we propose to generate surrogate-based explanations for individual predictions based on a sampling centered on particular place of the decision boundary, relevant for the prediction to be explained, rather than on the prediction itself as it is classically done.
We evaluate the novel approach compared to state-of-the-art methods and a straightforward improvement thereof on four UCI datasets.
\end{abstract}

\section{Introduction}

The task of explaining individual predictions made by a black-box classifier aims at providing to a human-user the rationale, or at least intuitions, about the factors leading to this prediction and, eventually, showing how the prediction can be altered by changing some of these factors~\cite{Doshi-Velez2017}.
While it is clear in the current literature that there are still no consensual definitions of ``explanations'' and ``interpretability'' for machine learning algorithms and their predictions, we can nevertheless state that there is a consensus to say that providing an explanation for an individual prediction relies on finding the features that actually impact the prediction.

To do so, several types of approaches have been proposed. In this paper we consider \emph{post-hoc} approaches, which are model-agnostic and classically applied to trained machine learning prediction models.
For instance, sensitivity analysis~\cite{simonyan2013deep,Adler2016a,Koh2017a} generates perturbations on the feature values of the instance of the prediction to explain in order to observe the consequences on the output of the black-box to outline the local behaviour of the black-box regarding its local decision boundary.
Surrogate models~\cite{Craven1996,Hara2016a}, on the other hand, train an interpretable model (e.g. linear regression or decision tree with low complexity) to mimic the black-box decisions in order to extract explanations from it. A particular case are the local surrogates~\cite{Ribeiro2016} which, in order to better locally approximate the black-box decision boundary, propose to train an interpretable model locally in the neighborhood of the instance whose prediction by the black-box is to explain.

In this paper, we refine the notion of \emph{locality}, the neighborhood on which a local surrogate is trained and show that it is not trivial to define the right neighborhood. We illustrate this issue using LIME~\cite{Ribeiro2016}: we show that choosing an adequate sampling strategy for generating the instances used to fit the surrogate model has a major impact on the quality of the approximation of the local black-box decision boundary and thus on the accuracy of the generated explanation. In particular, the effect of locally important features can be hidden by globally important ones. We show that this issue is not related to a simple parametrization to control the range of the sampled neighborhood: for instance, centering the sampling on the instance of the prediction to explain may not be the best location to approximate the black-box decision boundary.
To solve this issue, we propose a novel approach to sample the right neighborhood to fit local surrogate models. The intuition is the following: since a local surrogate aims at approximating the local black-box decision boundary that matters for the prediction to explain, this boundary should be sought first in order to sample instances in its neighborhood to fit the local surrogate.
This approach is experimented on both synthetic datasets and datasets from the UCI repository with significant improvement in the local fidelity of the surrogate.

The next section details the steps to generate interpretable local surrogates, with a focus on LIME that will be used in Section~\ref{section:issues_locality} to illustrate and analyze the locality issue of the sampling to fit local surrogates. Finally, in Section~\ref{sec:proposition} a novel approach is proposed to sample the right neighborhood to fit local surrogate models. The data and code used in this paper are available online\footnote{\href{https://github.com/axadil/locality-interpretable-surrogate}{https://github.com/axadil/locality-interpretable-surrogate}}.

\section{Interpretable Local Surrogates}

In this section we present the classical principles to generate interpretable local surrogates to extract explanations for individual predictions. Then we instantiate it with a current state-of-the-art and widely used in the industry approach, LIME~\cite{Ribeiro2016}. It is then also used in Section~\ref{section:issues_locality} to illustrate the \emph{locality} issue in the sampling step to approximate a black-box decision boundary with local surrogates.

\subsection{Principles}
\label{subsection:surrogate_principles}

We consider a black-box classifier $b:\mathcal{X}\rightarrow\mathcal{Y}$ whose predictions $b(x) \approx y \in \mathcal{Y}$ are not understandable by a human expert. Given an instance $x \in \mathcal{X}$ associated with the prediction $b(x)$, the task at hand is to provide human-interpretable explanations for the rationale behind $b(x)$ to approximate the local decision boundary of $b$ with a surrogate model $s_x$ from which explanations are extracted.

This process can be divided into three steps: (1) sample the feature space to generate a training set. Once the training set is obtained, (2) a surrogate can be fit using it in order to approximate the local decision boundary of $b$. Constraints can be introduced in the loss function of the surrogate to control the complexity of the generated explanations or to impact the locality of the surrogate (see Section~\ref{subsection:lime}). Finally, (3) explanations are extracted from the surrogate~$s_x$. The form of the explanations depends on the surrogate type and on the chosen interpretability approach. In this paper, we will not discuss the question of the human-interpretability of the extracted explanations. We rely on admitted interpretability of the surrogate models as presented in the literature~\cite{guidotti2018survey}.
These three steps rely on assumptions and heuristics. In the next sections we propose to discuss the ones made in the step that will be the focus of this work, the sampling.

Regarding the sampling step, to fit the local surrogate~$s_x$ to approximate the local black-box decision boundary, an adequate training set composed by a set of instances~$X_{s_x}$ associated with their labels~$Y_{s_x}$ is required.
Since our objective is to approximate~$b$ in a post-hoc set-up, the labels associated to~$X_{s_x}$ are obtained from the the black-box classifier: $Y_{s_x} = b(X_{s_x})$. A crucial question is how to generate the set of instances~$X_{s_x}$.
The instances in the training set $X$, already used to train $b$, could be used to fit the surrogate $s_x$. However, $X$ may not be accessible, in particular when considering model-agnostic explainer systems. Also, it has been shown in~\cite{Craven1996} that locally increasing the density of instances is beneficial for the surrogate accuracy. For these reasons, it has been proposed to draw a new set of instances from the feature space~$\mathcal{X}$.
To do so, a strategy for the sampling of the instances~$X_{s_x}$ must be defined, in particular the parameters of the distribution from which the instances are drawn (law, center and range of the distribution) to define the neighborhood in which the training set of the surrogate~$X_{s_x}$ should focus.
Different choices for these parameters directly impact the explanations, as the \emph{locality} of the sampled instances~$X_{s_x}$ leads to different approximations of the local black-box decision boundary by the surrogate~$s_x$.

Sections~\ref{section:issues_locality} and~\ref{sec:proposition} of this paper are respectively dedicated to a discussion on the impact of different strategies of sampling on the local fidelity of the surrogate~$s_x$ and a novel proposition to sample~$X_{s_x}$ in order to improve the local fidelity of the surrogate~$s_x$.

\subsection{LIME}
\label{subsection:lime}

LIME generates an explanation for an individual prediction~$b(x)$ by fitting a linear approximation of the local black-box decision boundary. LIME implements the process described above (Section~\ref{subsection:surrogate_principles}) as follows.
Regarding the sampling step~(1), a set of instances~$X_{s_x}$ is drawn following a normal distribution with the same mean and standard deviation as the original feature space~$\mathcal{X}$, independently from the instance~$x$ of the prediction~$b(x)$ to explain. For the labels $Y_{s_x} = b(X_{s_x})$, LIME works with the prediction probabilities returned by~$b$.
Regarding the surrogate fitting step~(2), the surrogate of LIME is trained to approximate locally the black-box decision boundary with a linear regression with regularization (ridge). To fit a \emph{local} surrogate centered on $x$, each instance of~$X_{s_x}$ is associated to a weight calculated as its distance to~$x$ using a kernel function (RBF kernel by default): instances closer to~$x$ are assigned a higher importance during the training.
Regarding the explanation generation step~(3), human-interpretable explanations for the prediction~$b(x)$ are generated by extracting the linear regression coefficients of the trained surrogate~$s_x$.

\section{A Discussion about \emph{Locality} Issues}
\label{section:issues_locality}

As discussed in the previous section, \emph{locality} can be enforced either in the sampling step or when fitting the surrogate. Ensuring that the surrogate is trained with the right locality is a major challenge to avoid the generation of inaccurate explanations.
This section is devoted to the concept of locality and aims at highlighting the importance of incorporating it adequately for all surrogate model approaches.

\subsection{Local vs. Global Features}

Generating a local explanation for an individual prediction relies on finding the features that have a local influence over this prediction.
In a supervised learning problem, we propose to make a distinction between two types of feature influences.
Some features are expected to have a \emph{global} influence when they impact predictions for instances over all the dataset and others to have a \emph{local} influence when they impact predictions for small areas of the whole feature space. For instance, in a trained decision tree, features used close to the root have a more global influence than the ones used only in nodes that are close to the leaves.

Black-box decision boundaries often have local nuances and non-linearities, where global explanations based on global feature influences are not accurate. A local surrogate model trained to generate explanations for an individual prediction should have the ability to catch these local nuances: the features that matter locally to approximate the local black-box boundaries and relevant to explain the individual prediction.
For instance, as explained earlier, LIME samples instances from the whole input space and weights them depending on their distance to the instance $x$ whose prediction we want to interpret. Our claim is that such an approach tends to hide the features with a local influence for the benefit of features with a global influence.

\subsection{Locality in the Sampling Step}

\begin{figure}[t]
\begin{center}
    \includegraphics[width=1\linewidth]{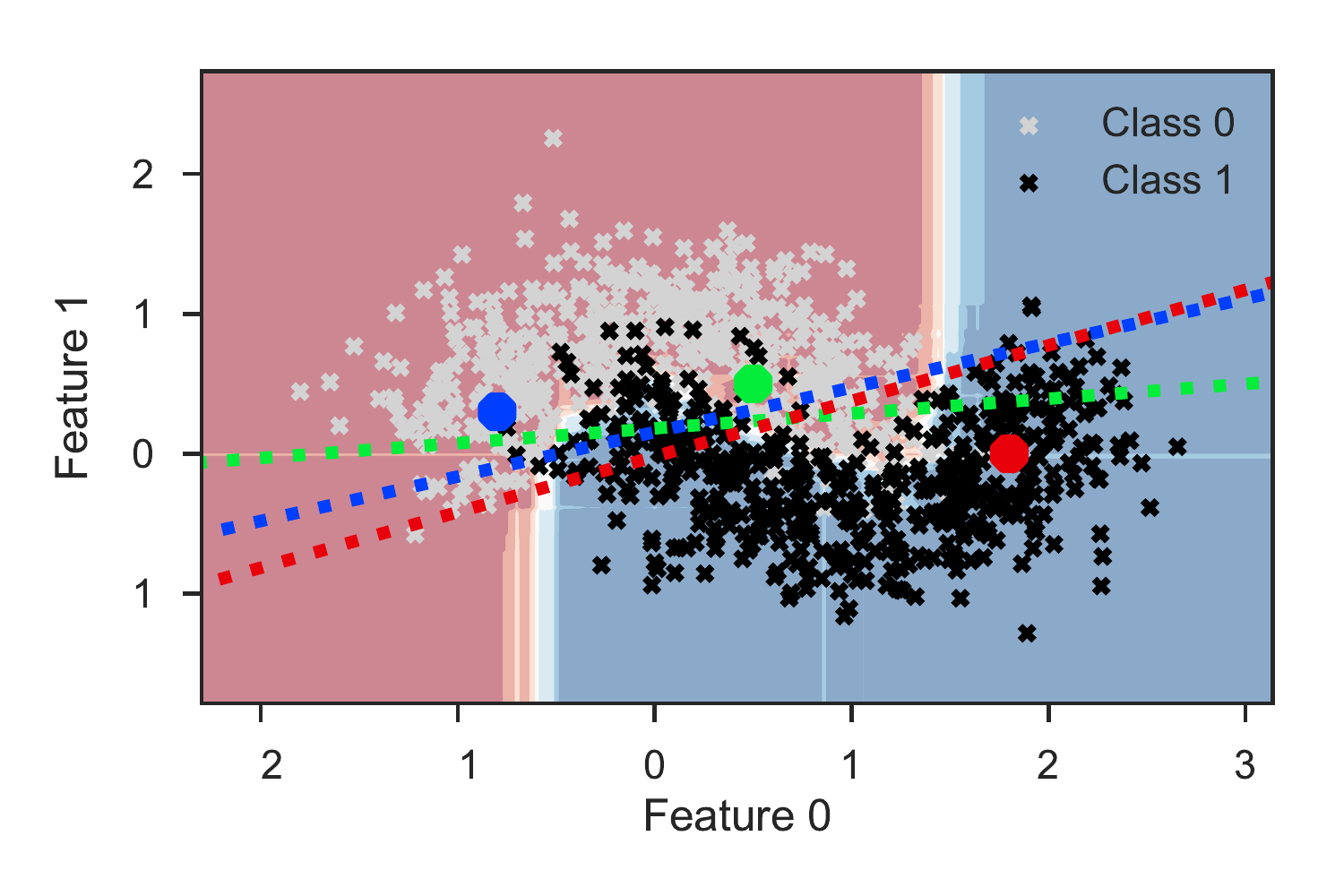}
    \caption{Local linear approximations (dashed colored lines) provided by LIME for 3 predictions (big color points) on half-moons dataset. The black-box decision boundaries are represented by the blue and red areas.}
    \label{fig:examples-LIME}
\end{center}
\end{figure}

To fit a local surrogate, the sampling step must handle the sampling of the right instances, what we call the \emph{sampling locality}, to feed and train a local surrogate. We use the decision boundary of the surrogates produced by LIME as an illustrative example.
It is easy to visualize the provided linear approximation of the decision boundary: LIME recreates a local classification decision boundary along the hyperplane defined by $s_x(a) = 0.5$, for $a \in \mathcal{X}$.
We apply LIME to a 2-dimensional half-moons dataset~(1000 instances, see Figure~\ref{fig:examples-LIME}), randomly split into a train set and a test set. A black-box classifier $b$, a Random Forest with the default \emph{scikit-learn} parameters in the present case, is trained on the train set resulting in a 0.93 AUC score on the test set.
In what follows, we use LIME as provided by the library developed by its authors~\cite{Ribeiro2016} with default parameters (if not mentioned otherwise). The library has been slightly modified to return the material needed to plot the LIME decision boundary.

Figure~\ref{fig:examples-LIME} illustrates the dataset and the decision border of $b$ as background (red vs. blue areas). In addition, for the 3 instances whose prediction is to explain (blue, green and red points, respectively $x_1$, $x_2$ and $x_3$), the boundaries learned by LIME are plotted as the blue, green and red dotted lines. 
We observe that the decision boundaries learned by LIME do not match the direction of the local decision boundary of the black-box classifier, where far more vertical borders could be expected. For~$x_2$~(green dot) for instance, looking at the shape of the closest decision boundary of~$b$, a negative slope could have been expected for the decision boundary learned by LIME rather than a positive one.
Another observation is that the slopes of the decision boundaries learned by LIME for these instances scattered across the dataset are similar, even though their respective local black-box decision boundaries are apparently different.
These results tend to show that LIME's decision boundaries sometimes approximate the global shape of the black-box decision boundary instead of its local one, resulting in local feature influences being mitigated in favour of global feature influence.

\subsection{Evaluation: a Numerical Criterion for Locality}

To get a quantitative evaluation of these observations, we propose a metric to assess the locality issue, and define the Local Fidelity as the fidelity of $s_x$ to $b$ within a neighborhood $\mathcal{V}_x$ around $x$:

\begin{equation}
    LocalFid(x, s_x) = Acc_{x_i \in \mathcal{V}_x}(b(x_i), s_x(x_i))
\end{equation}

where $Acc$ is a measure of accuracy, such as the AUC score, calculated over instances generated uniformly in $\mathcal{V}_x$. The surrogate $s_x$ is not affected by $\mathcal{V}_x$, since $\mathcal{V}_x$ is defined after $s_x$ and purely used for evaluation. We propose to define the neighborhood $\mathcal{V}_x$ of $x$ as a $l_2$-hypersphere of radius $r_{fid}$ centered on $x$. This intuitive definition of locality allows us to make the radius $r_{fid}$ of the fidelity hypersphere a proxy for the degree of locality considered. Since the radius value is heavily linked to the dimension and density of the input space $\mathcal{X}$, $r_{fid}$ will be expressed in the rest of the paper as a percentage of the maximum distance between the instances of the dataset and $x$, whose prediction is being interpreted. 

\begin{figure}[t]
\begin{center}
    \includegraphics[width=1\linewidth]{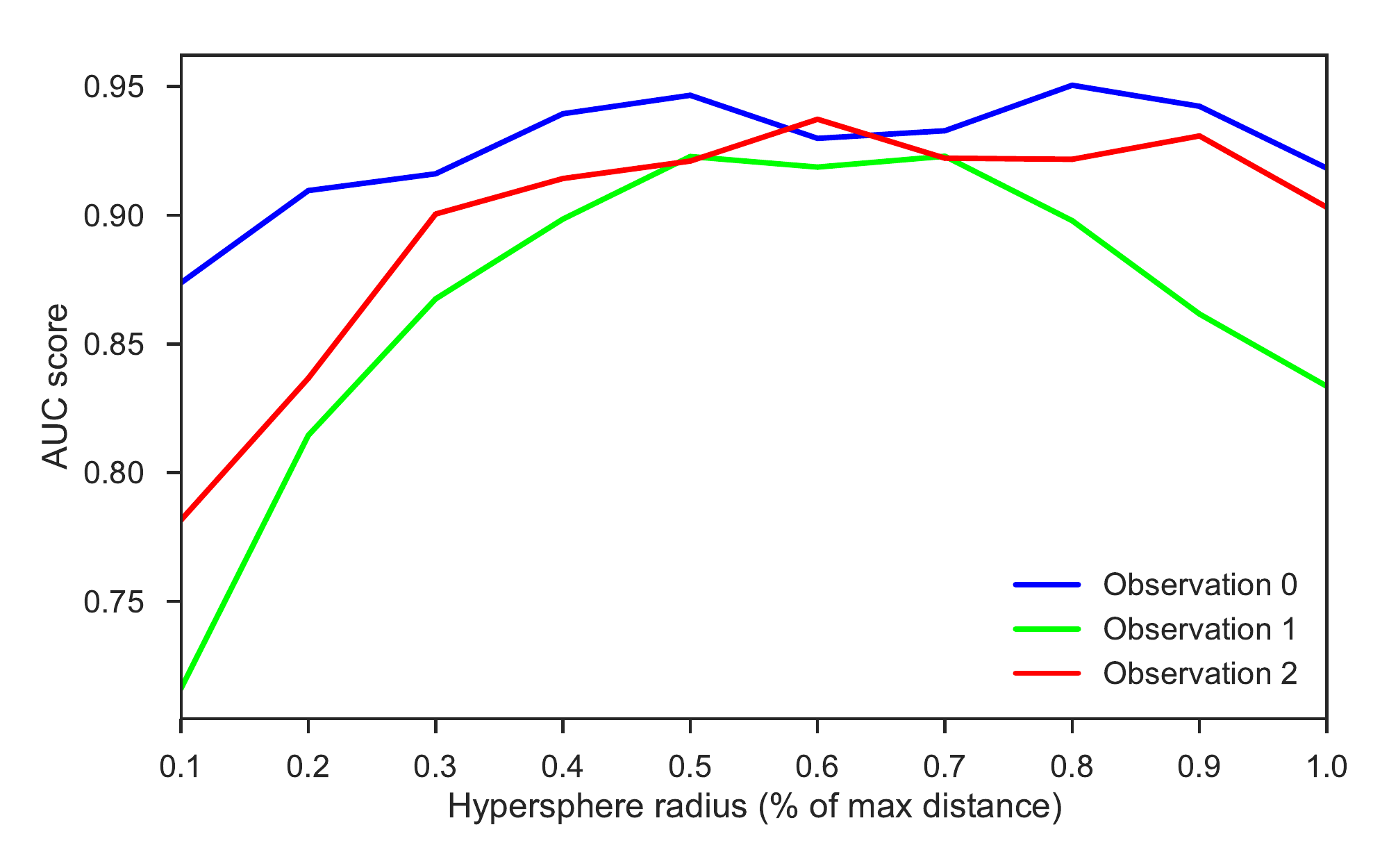}
    \caption{Local Fidelity for 3 predictions provided by LIME on half-moons dataset for increasing values of $r_{fid}$.}
    \label{fig:radius-LIME}
\end{center}
\end{figure}

Figure~\ref{fig:radius-LIME} shows the value of this criterion for LIME for the dataset and the 3 instances shown on Figure~\ref{fig:examples-LIME}, for different values of~$r_{fid}$. At a local scale, for low values of radius~$r_{fid}$, the fidelity is significantly worse than at global scale, for higher values of~$r_{fid}$. This confirms the previous observation: the approximation learned by the local surrogate is influenced by global features, which reduces the Local Fidelity at local scale of the local surrogate and thus the fidelity of the explanations it will generate for the individual prediction to explain.

\begin{figure}[t]
\begin{center}
    \includegraphics[width=1\linewidth]{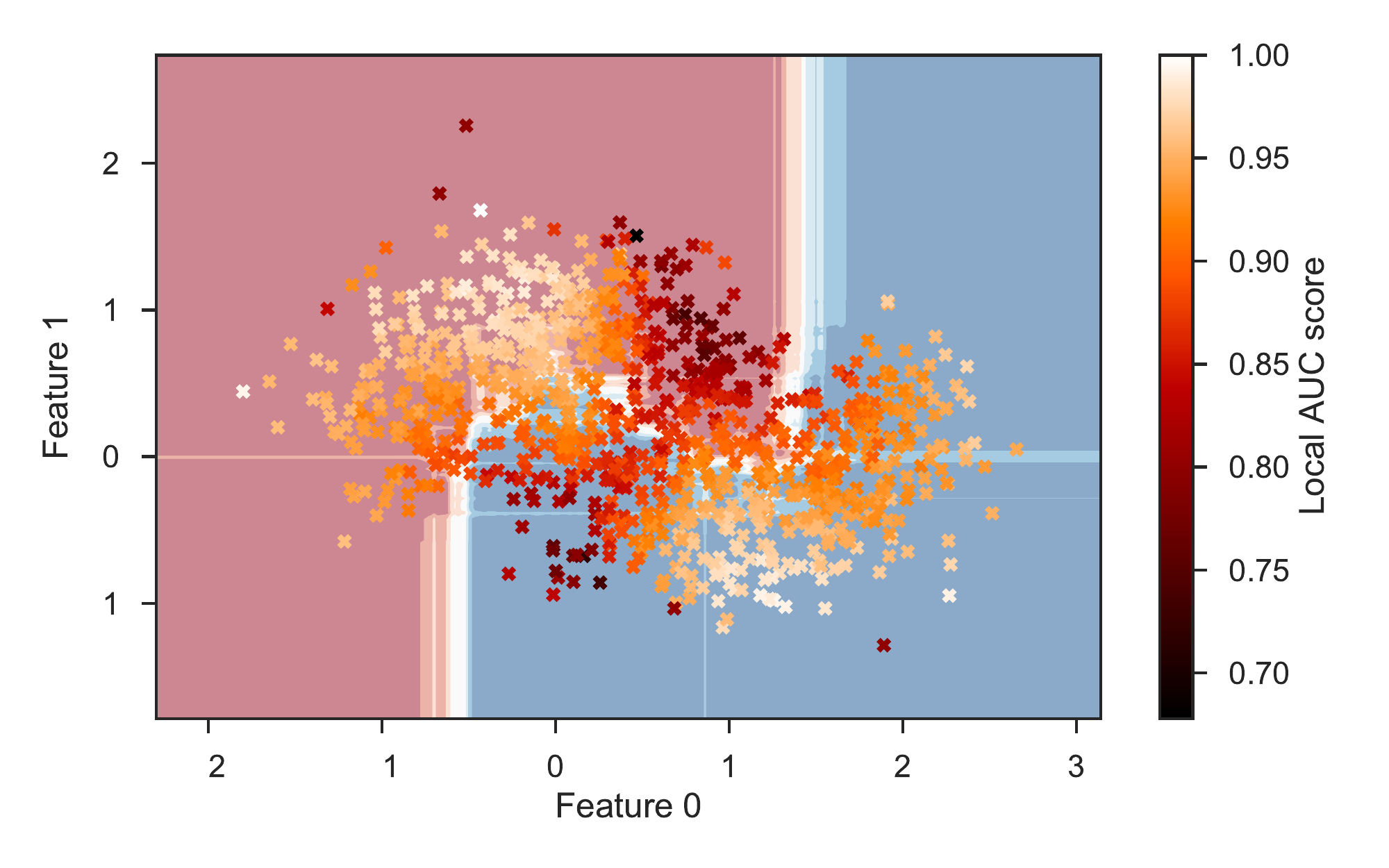}
    \caption{Visualization of the half-moons dataset and the Local Fidelity ($r_{fid}=0.3)$ of LIME for each instance of the dataset.
    }
    \label{fig:heatmap-LIME}
\end{center}
\end{figure}

In order to get insights about the quality of the local approximations of a surrogate model over a whole dataset, the definition of Local Fidelity can be extended to a set of instances $D=(x_i)_i$ by averaging $LocalFid$ over~$D$ in neighborhoods $\mathcal{V}_{x_i}$.
Figure~\ref{fig:heatmap-LIME} shows a heatmap where each point of the test set is colorized depending of the Local Fidelity of LIME for $r_{fid}=0.3$. We can observe that LIME has trouble approximating areas where the \emph{local} decision boundary of the black-box classifier~$b$  differs from the decision boundary approximating the whole dataset (ie. features with a local influence vs. features with a global influence).
Our hypothesis for the rationale behind this behaviour is that for a local surrogate to fit properly a local decision boundary of the black box, the local data sampled and used to fit the surrogate should be in the neighborhood of the decision boundary to approximate. Instead, LIME weights instances sampled over the whole dataset with a kernel function of the distance to the $x$ to explain, which does not seem to be sufficient.


\section{A New Local Surrogate Training Workflow}
\label{sec:proposition}

In order to overcome the issues highlighted in the previous section, a first idea consists in modifying the kernel width of LIME, resulting in a new surrogate model that we call LIME-K. Here, we propose a more fundamental modification of the sampling stage in the local surrogate training workflow, LS, as described in the following.

\subsection{Proposition : Local Surrogate (LS)}

The main idea of our proposition relies on the assumption that in order to approximate a local decision boundary, the data $X_{s_x}$ used for the training of the surrogate model should be sampled precisely around the decision boundary itself. Although the criteria to maximize, $LocalFid$, is calculated in an area centered around the instance~$x$, the final objective of $s_x$ remains to approximate the classification decision boundary of~$b$. Hence, sampling instances belonging to the two classes (which would not be guaranteed when sampling around~$x$) is important.

Given an individual prediction to explain $x$ and a black-box classifier $b$, our proposition for the sampling stage is as follows. First, the closest decision boundary of~$b$ is detected by looking for the closest instance $x_{border}$ such that $b(x_{border}) \neq b(x) $. This is done by using the Generation part of the $GrowingSpheres$ algorithm introduced in \cite{laugel2017inverse}. The overall principle is to generate instances in a hypersphere of growing radius centered on~$x$ until it crosses the decision boundary of $b$.
Once~$x_{border}$ is found, training instances are sampled uniformly in a hypersphere~$\mathcal{S}$ of radius~$r_{s_x}$ centered on~$x_{border}$ so that the shape of this closest decision border can be efficiently learnt: $X_{s_x}\sim \mathcal{U}_{\mathcal{S}(x_{border}, r_{s_x})}$. The outline of the method is detailed in Algorithm \ref{algo}.

Using adequate hyperparameters for $GrowingSpheres$ (such as the number of instances generated at each step) guarantees to detect one of the closest points of the decision border of $b$. 
Thus, $LocalSurrogate$ performs an approximation of the local decision boundary of $b$. The radius $r_{s_x}$ used to define the training subset of $s_x$ defines how local this approximation is.

\begin{algorithm}[t]
\begin{algorithmic}
   \STATE {\bfseries Input:} $x \in \mathcal{X}$, $b:\mathcal{X}\rightarrow\mathcal{Y}$, $r_{s_x}$, $N$
   \STATE $x_{border} \leftarrow GrowingSpheres(b, x)$
   \STATE $X_{s_x} \leftarrow$ Draw uniformly $N$ instances in an hypersphere of radius $r_{s_x}$ centered around $x_{border}$
   \STATE $Y_{s_x} \leftarrow b(X_{s_x})$
   \STATE Train $s_x$ on ($X_{s_x}, Y_{s_x}$)
   \STATE {\bfseries Return:} $s_x$
\end{algorithmic}
\caption{Outline of LocalSurrogate algorithm}
\label{algo}
\end{algorithm}

\subsection{Experimental Results}

The experimentation is conducted on the previous artificial half-moons dataset and on 4 UCI datasets. 

\subsubsection{Competitors}

The proposed algorithm $LS$ is evaluated in comparison with LIME using the criterion presented in Section~\ref{section:issues_locality}. As shown in \cite{Ribeiro2016} and discussed in Section~\ref{subsection:lime}, LIME ensures some level of locality by using a kernel to weight the sampled instances depending on their distance to the instance~$x$ whose prediction we want to to interpret. The width of this kernel could be set by the user to constraint locality in a more or less aggressive fashion. 
In this context, the proposed Local Surrogate approach is compared to LIME with default parameters (kernel width equals $0.75 \; \sqrt{dimension(\mathcal{X})}$) and to LIME with a reduced kernel width (that is optimized for each dataset based on performance) that we call LIME-K.

\subsubsection{Half-moons dataset}

\begin{figure}[t]
\begin{center}
    \includegraphics[width=1\linewidth]{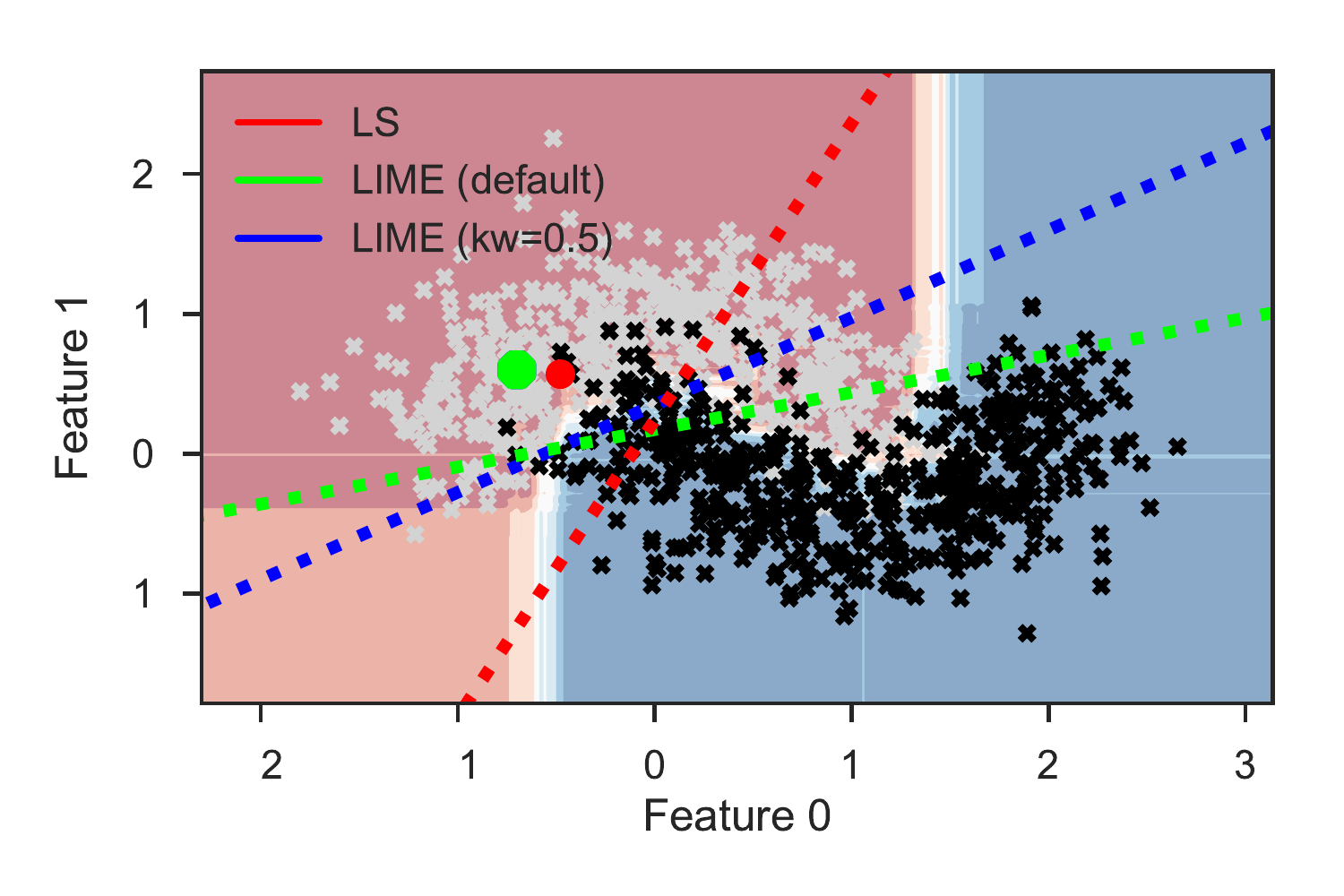}
    \caption{Example of the linear approximations performed by LIME (with default kernel width), LIME-K (with reduced kernel width) and the proposed Local Surrogate for a randomly picked instance from the half-moons dataset (green dot $x$). The red dot corresponds to the closest instance from the other class~$x_{border}$ found with the $GrowingSpheres$ algorithm.}
    \label{fig:examples-prop}
\end{center}
\end{figure}

\begin{figure}[t]
\begin{center}
    \includegraphics[width=1\linewidth]{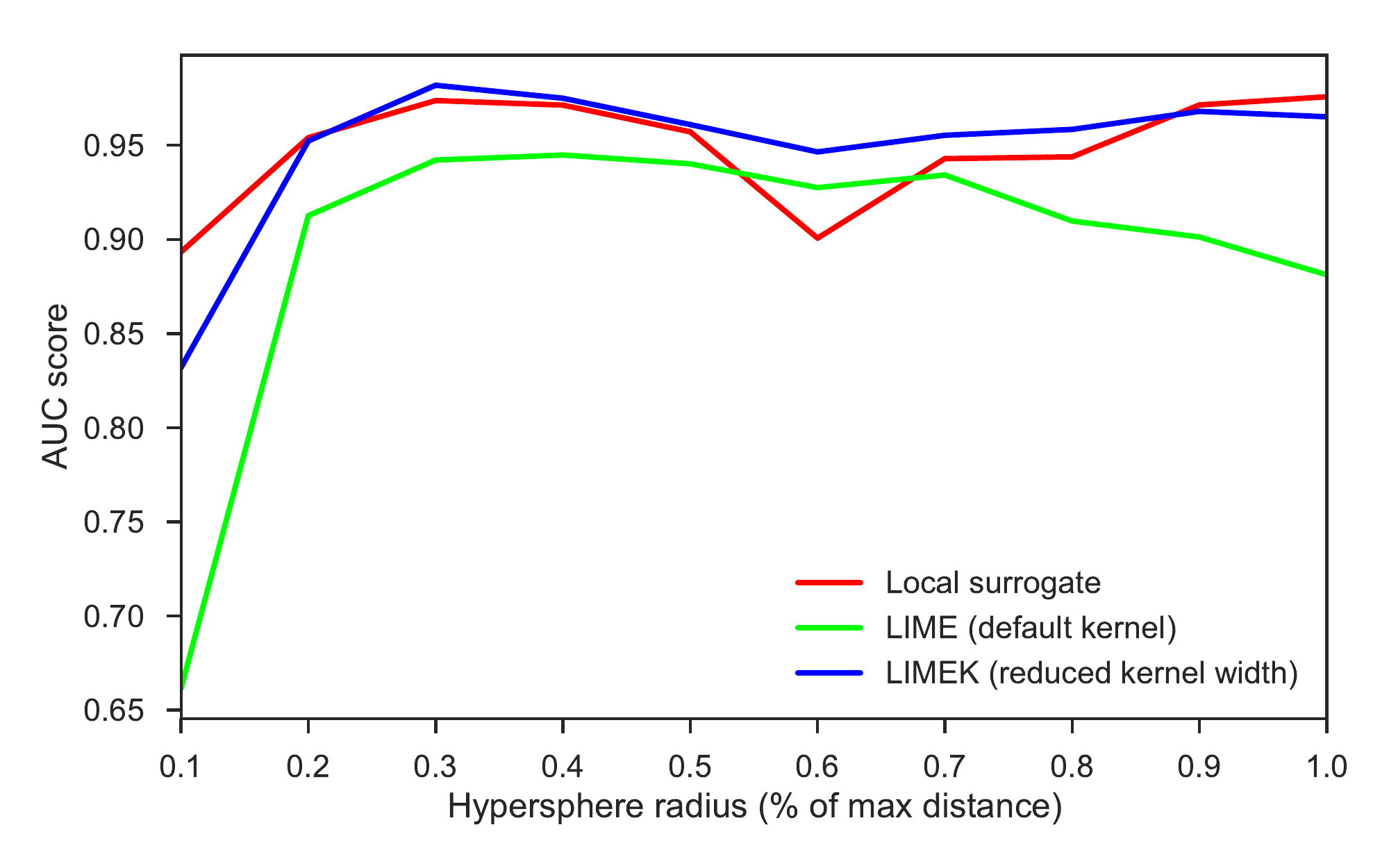}
    \caption{Local Fidelity for an instance of the half-moons dataset for several values of~$r_{fid}$.}
    \label{fig:radius-prop}
\end{center}
\end{figure}

We analyze the efficiency of the proposed approach at bringing locality in the same setup as before~(Random Forest trained on half-moons) to facilitate the visualization of locality effects being learned by the local surrogate model. Figure~\ref{fig:examples-prop} shows the decision boundaries of LIME (default kernel width $\sigma=0.75 \; \sqrt{dimension(\mathcal{X})}$), LIME with a reduced kernel width parameter ($\sigma=0.5$) and ther proposition $LS$ ($r_{s_x}=0.3$) for a randomly picked instance of the dataset. As shown previously, the decision boundary learned by LIME (in green) is very horizontal, as an approximation of the global model would be. Even though reducing the kernel width helps in making the learned decision boundary more local (blue line), it is not enough to properly approximate the local border of the black-box classifier $b$. In comparison, $LS$ seems to be approximating a much more local border direction (red line), as its slope is much more vertical, as expected for this instance (green dot), giving us a more satisfying explanation.

These results can be further observed when looking at the $LocalFid$ scores of these learned decision boundaries for a given instance in Figure~\ref{fig:radius-prop}. The proposed Local Surrogate approach achieves higher Local Fidelity than LIME, even with reduced kernel width. This can be generalized to the whole test set, as shown by the differences in average Local Fidelity ($r_{fid}=0.2$) in Table~\ref{tab:avgLocalAcc-uci}.
This tends to confirm our initial assumption that even if weighting helps to integrate locality, the global sampling performed by LIME tends to mitigate the local feature effects in favor of the global ones.

\subsubsection{UCI datasets}

We apply the same experimental protocol to 4 classification datasets from the UCI repository (Breast Cancer, Default of Credit Card Clients, Online News Popularity and Tennis Major Tournament Match Statistics~\cite{Dua:2017}). For each dataset, we keep the numerical attributes, split it between train and test sets (test size set to $20\%$) and train a Random Forest classifier with 200 trees and default parameters from the scikit-learn package. Once again, it is important to note that the considered local surrogates approaches are model-agnostic, and therefore the choice of the classifier does not matter.

We then calculate the average Local Fidelity ($r_{fid}=0.05$) of LIME (with default kernel width), LIME-K (LIME with reduced kernel width) and the proposed Local Surrogate approach ($r_{s_x}=0.3$) in the whole test dataset. The results are shown in Table~\ref{tab:avgLocalAcc-uci}.

The average Local Fidelity of the proposed Local Surrogate approach is significantly higher~(between $+0.08$ and $+0.18$ in AUC score across all datasets) than the one obtained with LIME and LIME-K.
Despite an optimized kernel width, LIME thus fails at properly approximating the black-box classifier locally consistently over the whole dataset, resulting in high standard deviation values. On the other hand, LS achieves better Local Fidelity across all datasets with lower standard deviation, thus providing more accurate local explanations for the predictions made by $b$. 

\begin{table}[t]
    \centering
    \begin{tabularx}{\linewidth}{X|XXX}
      \hline
      Dataset & LIME & LIME-K & LS \\
      \hline
      \hline
       \nicefrac{1}{2} moons & 0.89 (0.07) & 0.96 (0.06) & \textbf{0.97 (0.03)} \\
      cancer & 0.86 (0.07) & 0.87 (0.07) & \textbf{0.96 (0.02)} \\
      credit & 0.67 (0.21) & 0.70 (0.18) & \textbf{0.85 (0.12)} \\
      news & 0.64 (0.10) & 0.67 (0.10) & \textbf{0.79 (0.07)} \\
      tennis & 0.85 (0.12) & 0.83 (0.13) & \textbf{0.98 (0.02)}\\
      \hline
    \end{tabularx}
    \caption{Average and standard deviation Local Fidelity scores ($r_{fid}=0.05$) for LIME (default), LIME-K and $LS$ (our proposition) over the half-moons dataset and 4 UCI datasets}
    \label{tab:avgLocalAcc-uci}
\end{table}

\section*{Conclusion}

In this work, we highlighted the importance of the locality when training local surrogates to provide human-interpretable explanation for a prediction made by a black-box classifier. Features with a local influence can be easily mitigated by the effect of features with a global influence. The challenge lies in defining a relevant sampling method to generate a dataset that allows the surrogate to approximate accurately the black-box decision boundary.
We contributed in that direction with the proposition of the Local Surrogate approach. It proposes to center the generation of the surrogate's training set on the black-box decision boundary and in its immediate neighborhood to ensure a better approximation.
We showed improvements provided by our proposition on both toy and UCI datasets in terms of local surrogate fidelity.

Ongoing works focus on the human-interpretability benefits of $LS$, in collaboration with experts from the insurance industry. We also plan to ease the setup of the hyper-parameters of the proposed metrics and approach so they are easier to use and take into account situations were they would lead to an inadequate level of locality (e.g. when using a radius~$r_{fid}$ too small to cover the decision border) and thus less relevant explanations.

\section*{Aknowledgements}
This work has been done as part of the Joint Research Initiative (JRI) project
”Interpretability for human-friendly machine learning models” funded by the
AXA Reseach Fund.

\bibliography{mendeley}
\bibliographystyle{icml2018}

\end{document}